\documentclass[10pt,twocolumn,letterpaper]{article}

\usepackage{cvpr}              

 \usepackage{tabularx}
 \usepackage{multirow}

\definecolor{cvprblue}{rgb}{0.21,0.49,0.74}
\usepackage[pagebackref,breaklinks,colorlinks,allcolors=cvprblue]{hyperref}

\def\paperID{31528} 
\def\confName{CVPR}
\def\confYear{2026}

\title{Neuro-Cognitive Reward Modeling for Human-Centered Autonomous Vehicle Control}

\author{Zhuoli Zhuang, Yu-Cheng Chang, Yu-Kai Wang, Thomas Do, Chin-Teng Lin
\\
University of Technology Sydney\\
{\tt\small \{zhuoli.zhuang, fred.chang, yukai.wang, thomas.do, chin-teng.lin\}@uts.edu.au}
}

\begin{document}
\maketitle
\thispagestyle{empty}
\begingroup
\renewcommand\thefootnote{}
\footnotetext{Accepted to the IEEE/CVF Conference on Computer Vision and Pattern Recognition (CVPR) 2026 Main Track.}
\endgroup
\begin{abstract}
Recent advancements in computer vision have accelerated the development of autonomous driving. Despite these advancements, training machines to drive in a way that aligns with human expectations remains a significant challenge. Human factors are still essential, as humans possess a sophisticated cognitive system capable of rapidly interpreting scene information and making accurate decisions. Aligning machine with human intent has been explored with Reinforcement Learning with Human Feedback (RLHF). Conventional RLHF methods rely on collecting human preference data by manually ranking generated outputs, which is time-consuming and indirect. In this work, we propose an electroencephalography (EEG)-guided decision-making framework to incorporate human cognitive insights without behaviour response interruption into reinforcement learning (RL) for autonomous driving. We collected EEG signals from 20 participants in a realistic driving simulator and analyzed event-related potentials (ERP) in response to sudden environmental changes. Our proposed framework employs a neural network to predict the strength of ERP based on the cognitive information from visual scene information. Moreover, we explore the integration of such cognitive information into the reward signal of the RL algorithm. Experimental results show that our framework can improve the collision avoidance ability of the RL algorithm, highlighting the potential of neuro-cognitive feedback in enhancing autonomous driving systems. Our project page is: https://alex95gogo.github.io/Cognitive-Reward/.
\end{abstract}    
\section{Introduction}
\label{sec:intro}

Autonomous intelligent vehicles (AIVs) have greatly benefited from deep learning in recent years. Thanks to deep neural networks' strong feature extraction capabilities, end-to-end autonomous driving (E2E-AD) has emerged as a promising approach. This approach directly maps raw sensor inputs, such as camera images, to control signals. In the CARLA driving simulator \cite{dosovitskiy2017carla}, several E2E-AD models \cite{shao2023reasonnet, shao2023safety, wu2022trajectory, duan2024enhancing} have achieved driving performance comparable to that of rule-based expert systems, as demonstrated in the CARLA leaderboard 1.0 \cite{carla_leaderboard}.
Despite these advancements, recent studies have shown that E2E-AD models primarily replicate expert trajectories rather than exhibiting human-like reasoning abilities. This limitation becomes evident in out-of-distribution scenarios, where models may fail due to their inability to generalize beyond their training data. For instance, \cite{jaeger2023hidden} highlights how E2E-AD models often cut turns incorrectly to reach a target point, revealing a lack of true decision-making skills. In addition, a recent study \cite{jia2025bench2drive} found that top-performing E2E-AD models from the CARLA leaderboard struggle in interactive driving scenarios, such as emergency braking, likely due to the lack of explicit guidance on decision-making and interaction dynamics. 
A possible cause of these failures may be the limitation of imitation learning (IL), a predominant training paradigm in E2E-AD. IL-based methods rely on direct supervision of control signals or trajectory replication. However, IL
faces a notable limitation of the distribution shift problem~\cite{chang2021mitigating}, which means IL models do not learn from sufficient unforeseen failure cases as they learn a specific distribution from an expert~\cite{toromanoff2020end, filos2020can, zheng2022imitation, yu2023offline}.

Unlike IL, reinforcement learning (RL) has a learning paradigm from trial and error, dynamically adapting to new environments based on a reward signal \cite{mnih2013playing}, and therefore has the potential to mitigate the distribution shift problem. However, without explicit alignment mechanisms, neither RL nor IL guarantees alignment with human expectations. To address this gap, reinforcement learning with human feedback (RLHF) has emerged as a powerful approach to align AI behavior with human preferences \cite{christiano2017deep}. Traditional RLHF typically involves human annotators ranking AI-generated outputs or comparing trajectory segments based on replayed demonstrations. While effective, this process is time-consuming and relies on explicit feedback, which may not fully capture users' cognitive responses. In this work, we introduce a novel method for collecting human preference signals by leveraging natural neural responses recorded via electroencephalography (EEG), allowing for a more intuitive and efficient alignment of AI decision-making with human cognition.

 EEG is a non-invasive method for monitoring neural activity, such as event-related potentials (ERPs), which are time-locked responses to specific cognitive or sensory events. One of the most prominent ERP components is the P3, a positive peak occurring 300-500 ms after stimulus onset. Research has shown that an unexpected, unusual, or surprising stimulus, even if task-irrelevant, can elicit a frontal P3-like response within an attended sequence~\cite{courchesne1975stimulus, soltani2000neural, polich2003p3a}. 
The human brain contains billions of neurons, and the firing of these neurons produces electrical responses. ERPs originate as postsynaptic potentials, which occur when similarly oriented neurons fire synchronously in response to visual or auditory stimuli~\cite{luck2014introduction}. Since the discovery that brain wave activity peaks approximately 300 milliseconds after exposure to an unpredictable stimulus~\cite{sutton1965evoked}, ERPs have been extensively studied in neuroscience as reliable biomarkers due to their millisecond-level temporal resolution~\cite{luck1994electrophysiological,isreal1980p300,heinze1994combined}.
A classic experimental design for eliciting ERPs is the ``oddball paradigm", wherein the ERP amplitude is typically larger in response to relatively infrequent and unexpected events~\cite{luck1994electrophysiological}. Furthermore, ERP amplitudes have been shown to increase with task difficulty~\cite{isreal1980p300}. ERPs also provide complementary information to eye-tracking data. For instance, covert attention, the ability to focus on objects outside the fovea, cannot be detected with eye-tracking but is identifiable through ERP analysis~\cite{heinze1994combined}.

This study leverages the advantages of ERPs, specifically their reliable millisecond-level representation of natural brain responses. It extends classical ERP analysis by exploring the potential of using ERP data to train RL models for AIV tasks.
In this study, we collect EEG data from 20 participants as they actively drive in a realistic driving simulator. By analyzing their ERP responses, we find a correlation between ERP amplitude and driver reaction time measurements. To leverage this neuro-cognitive insight, we design and train a human cognitive reward model to predict ERP strength from scene images, allowing us to integrate this prediction into the reward signal for reinforcement learning. Rather than traditional RL with EEG feedback, which requires the collection of EEG data during inference \cite{wang2022implicit, wang2022error, wang2020maximizing, xu2021accelerating}, this approach does not require human data during inference, and is therefore more scalable. This approach enables RL agents to incorporate human cognitive feedback, improving their decision-making in autonomous driving.


Our main contributions are summarized as follows:

\begin{itemize}
    \item We present a novel dataset collected using a realistic VR driving simulator, which captures drivers' behavioral data, eye-tracking data, EEG signals, and corresponding scene images.
    
    \item We develop a human cognitive reward model that predicts EEG features from scene images.
    
    \item We propose an RLHF framework that integrates reward signals from the cognitive reward model and demonstrate that it can enhance the performance in two complex driving scenarios.
\end{itemize}

\begin{figure*}
\centerline{\includegraphics[width=\linewidth]{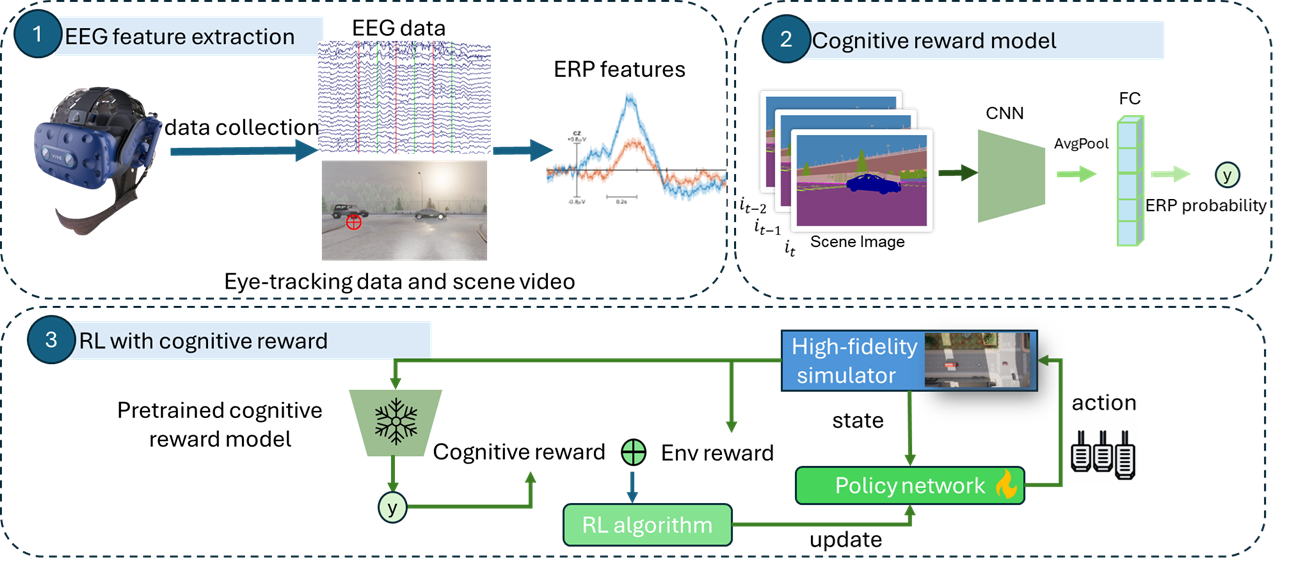}}
\caption{The framework of the human cognitive reward model. First, raw EEG data are preprocessed, and the most prominent feature, which is the ERP, is extracted. Second, an EEG feature prediction model predicts the ERP from the scene images. Lastly, the prediction probability is used as part of the reward of reinforcement learning for autonomous driving tasks.}
\label{fig:framework}
\end{figure*}

\section{Related Work}
\label{sec:related_work}

\subsection{RL with human feedback}
RL has demonstrated remarkable success in complex tasks, such as playing Atari games~\cite{mnih2013playing} and predicting protein structures~\cite{jumper2021highly}. However, RL typically requires carefully crafted reward functions, and the reward signals are often sparse, which hinders RL's development and application. In addition, RL may not align with human expectations because it optimizes behavior purely based on predefined reward functions, which often fail to capture the complexity of human values and expectations. To address this challenge, RLHF~\cite{christiano2017deep} was introduced. RLHF trains a reward model by learning from human preferences, typically by having humans rank or compare model outputs, and then uses this model to guide the RL agent.
RLHF has since become a paradigm for fine-tuning large language models (LLMs), aligning them more closely with human preferences~\cite{ouyang2022training}. It has been successfully applied in state-of-the-art LLMs such as ChatGPT~\cite{achiam2023gpt}, LLAMA~\cite{dubey2024llama}, and Gemini~\cite{team2024gemini}. 
This work builds on these advances by exploring the application of brain modeling methods in reward modeling for AIV tasks, which models a more intuitive or natural process, and extends the human-feedback-driven reinforcement learning.

\subsection{RL with EEG feedback}
Human physiological signals offer a more natural source for guiding RL models. For instance, leveraging neural networks to approximate eye-tracking features has been explored to enhance RL policy models~\cite{lopez2024seeing, zhuang2025aegis}. Yet, brain signals, a more direct source of human feedback, remain largely unexplored. One such signal is the error-related potential (ErrP), an ERP involuntarily elicited when a human perceives unexpected errors in an environment \cite{gehring1993neural, van2004modulation}. ErrP has been explored in various human-machine shared autonomy tasks.
For example, ErrP has been utilized to provide implicit feedback to RL agents when a human observes an error, demonstrating effectiveness in grid-world navigation tasks \cite{wang2022implicit, wang2022error}. Additionally, ErrP has been employed to train human policy models, which offer feedback to RL agents when the robot's uncertainty is high \cite{wang2020maximizing}. Similarly, ErrP has been applied for reward shaping in human-in-the-loop RL to accelerate training, proving beneficial in navigation games \cite{xu2021accelerating}.
Building on these theoretical groundings, this work leverages ErrP to guide RL agents. Prior approaches face two main limitations: they require real-time EEG data collection during inference, and ErrP signals occur only after an error event, which may introduce latency in corrective actions. To address these challenges, we propose a novel method that predicts EEG features directly from image data. This approach enables scalable and efficient training by eliminating the need for continuous EEG recordings, making RLHF more practical and accessible.

\subsection{Predicting brain response from images}

Advances in deep learning have made it possible to model brain responses to visual stimuli. For example, magnetoencephalography (MEG), electroencephalography (EEG), and functional magnetic resonance imaging (fMRI) activities have been modeled using latent representations of visual stimuli~\cite{cichy2017dynamics, gifford2022large, yang2024brain}. For instance, linear encoding models could predict EEG responses to arbitrary images, highlighting the possibility of developing models that bridge visual inputs and brain activity~\cite{gifford2022large}. In addition, a tool called FactorTopy is used to map features from pre-trained deep neural networks onto brain activity patterns measured via fMRI~\cite{yang2024brain}. In line with this, the researchers found that deep neural networks could effectively model the temporal progression of scene processing in the brain in an EEG study~\cite{cichy2017dynamics}. This work aims to build on these advances by introducing a framework that predicts brain responses to images for reward modeling. 
\section{Dataset}
\subsection{Data collection setup}

This study recruited 32 healthy participants (mean age: $25.6 \pm 4.6$ years), all of whom were rigorously screened to ensure optimal health status. Data from 20 participants were included in the final analysis, while the remaining 12 were excluded due to VR-induced motion sickness or insufficient engagement in the driving tasks. Participants exhibited normal or corrected-to-normal auditory and visual acuity, with pre-participation evaluations confirming the absence of neurological or psychiatric conditions, as well as current alcohol or substance misuse. These criteria were established to maintain the integrity and reliability of the data collected for the study. Participants were compensated for their involvement at a rate exceeding the national minimum wage. The participants had an average duration of driver’s license possession of $4.9  \pm 3.9$ years. Before the experiment, they completed the Driver Skill Inventory (DSI) questionnaire \cite{lajunen1995driving}, achieving a mean perceptual-motor skills score of $4.1 \pm 0.3$ and a mean safety skills score of $3.7 \pm 0.3$ on a scale ranging from 1 to 5. Informed written consent was obtained from all participants before their involvement in the study. The research protocol was reviewed and approved by the University's Human Research Ethics Committee, ensuring compliance with ethical standards for studies involving human subjects.
We used an HTC Vive Pro Eye VR headset ($2\times1440\times1600$ resolution, 90 Hz refresh rate, 3D spatial sound) and the CARLA simulator to develop the scenarios, and Logitech
G923 Racing Wheel and Pedal to collect active driving data. The EEG signals were recorded by the Synamps2 system 
using 64 channels with Ag/AgCl electrodes according to the international 10–20 system, following the setting in~\cite{zhuang2025aegis}. The sampling rate of the raw EEG data is 1000 Hz, and the impedances of the electrodes were kept below $15k \Omega$ during the experiment. 
Each scenario was conducted in a separate session lasting approximately 6 minutes to minimize participant fatigue. Participants were allowed to take sufficient rest between sessions. Participants were instructed to perform the driving tasks to the best of their ability, with the goal of reaching the destination as fast as possible while avoiding collisions.
To the best of our knowledge, our dataset is the first dataset to include EEG, gaze, and active control data from drivers.

\begin{table*}[h]

\centering
\begin{tabularx}{\textwidth}{
        @{}l  
        c     
        @{\hspace{5pt}}c  
        c     
        c     
        @{\hspace{5pt}}c  
        @{\hspace{5pt}}c  
        @{\hspace{5pt}}c  
        @{\hspace{5pt}}c  
        @{\hspace{5pt}}c  
    }
\toprule
Dataset  & Active ctlr. & Vehicle data & Camera   & Hazards & View & \#subjects & \# Frames & EEG & Gaze \\ 
\midrule
Ours & + & + & $S^{rgb,depth,sem}$  & + & VR & 20 & 720K & + & + \\ 
Collision Threat~\cite{markkula2021accumulation} & + & - & - & - & screen & 25 & - & + & -\\
Sustained-attention~\cite{cao2019multi} & + & + & - & + & screen & 27 & - & + & - \\
MPDB~\cite{tao2024multimodal} & + & - & - & + & screen & 35 & - & + & +\\
DrFixD(night) \cite{deng2023driving} & - & - & $S^{rgb}$ & - & screen & 31 & 67K* & - & + \\ 
LBW \cite{kasahara2022look} & - & - & $S^{rgb,depth}$ & - & screen & 28 & 123K* & - & +\\ 
CoCAtt \cite{shen2022cocatt}  & + & + & $S^{rgb}$ & - & screen & 11 & 17K* & - & +\\ 
MAAD \cite{gopinath2021maad} & - & - & $S^{rgb}$  & - & screen & 23 & 60K & - & +\\
TrafficGaze \cite{deng2019drivers} & - & - & $S^{rgb}$ & - & screen & 28 & 77K* & - & +\\ 
DADA-2000 \cite{fang2019dada} & - & - & $S^{rgb}$ & + & screen & 20 & 658K  & - & +\\ 
DR(eye)VE \cite{alletto2016dr} & + & - & $S^{rgb}$ & - & on-road & 8 & 555K & - & +\\ 
BDD-A \cite{xia2019predicting} & - & + & $S^{rgb}$ & + & screen & 45 & 378K* & - & +\\ 
C42CN \cite{taamneh2017multimodal} & + & - & $S^{rgb}$ & + & screen & 68 & - & - & +\\ 
TETD \cite{deng2016does}  & - & + & $S^{rgb}$ & - & screen & 20 & 100 & - & +\\
3DDS \cite{borji2011computational} & + & - & $S^{rgb}$ & - & screen & 10 & 192K & - & +\\ \bottomrule
\end{tabularx}

\label{table:supp_dataset}

\caption{Comparison between our dataset and other driving datasets. Ours contains the largest number of frames and is collected using a realistic VR driving simulator. Abbreviations: Camera – S: scene-facing camera, RGB: 3-channel image, sem: semantic segmentation mask. Frame counts marked with * are estimates based on video duration and frame rate.}

\end{table*}

\subsection{Scenario Design}

\textbf{Emergency Braking Scenario:} Inspired by \cite{jia2025bench2drive}, we developed a dynamic and interactive emergency braking scenario that presents a significant challenge to current autonomous driving models. This scenario is particularly designed to induce failure cases, thereby testing the robustness of both human and AI driving systems.
Additionally, our emergency-braking task is modeled after the adaptive cruise control (ACC) paradigm, as described by \cite{wang2022velocity}. In this task, the ego vehicle is instructed to closely follow a lead vehicle in the same lane, maintaining a position that prevents trailing vehicles from overtaking. Participants were informed that a trailing vehicle was closely following them, creating a sense of urgency to maintain speed and avoid being overtaken. The lead vehicle travels at speeds of up to 8 m/s and may brake unexpectedly, requiring the ego vehicle to respond promptly to avoid collisions while maintaining effective following distance.
This scenario presents substantial cognitive demands, as it requires continuous attention and real-time adjustments over extended periods to track the behavior of the lead vehicle accurately. To prevent participants from anticipating the event, the lead vehicle braked at a random interval of 4–7 seconds after the previous event.
Participants completed two sessions of active driving under different lighting conditions, one during the day and one at night. 
In addition, they participated in two control sessions where an AI system managed acceleration and braking, and they only needed to control the steering wheel, providing a baseline condition for ERP analysis.
\\
\textbf{Left Turn Scenario:}
The left turn scenario followed the design in \cite{wu2022prioritized}. Here, the ego vehicle is required to execute a left turn at an intersection while avoiding collisions with oncoming vehicles from the right. These oncoming vehicles approach at variable speeds ranging from 3m/s to 5m/s and do not yield to the ego vehicle. The ego vehicle is afforded the autonomy to merge into the traffic flow at any opportune moment to reach the designated goal point as swiftly as possible. To test the generalization of our algorithm, we trained the models in Town01 and tested the models in Town05.

\begin{figure}[htb!]
\centering
  \includegraphics[width=0.9\linewidth]{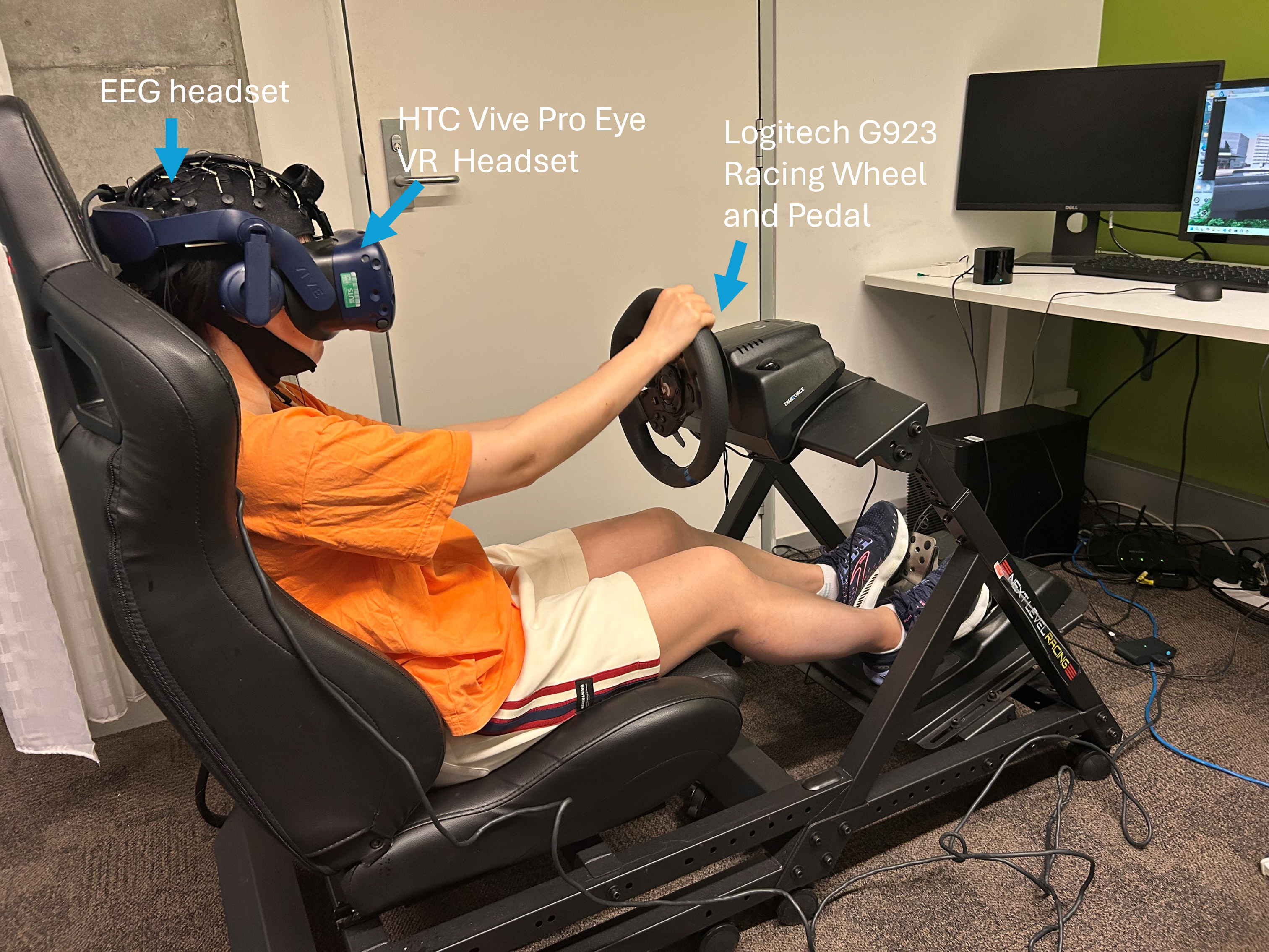}%
  \label{fig:sub-first}
\hfill
  \includegraphics[width=0.98\linewidth]{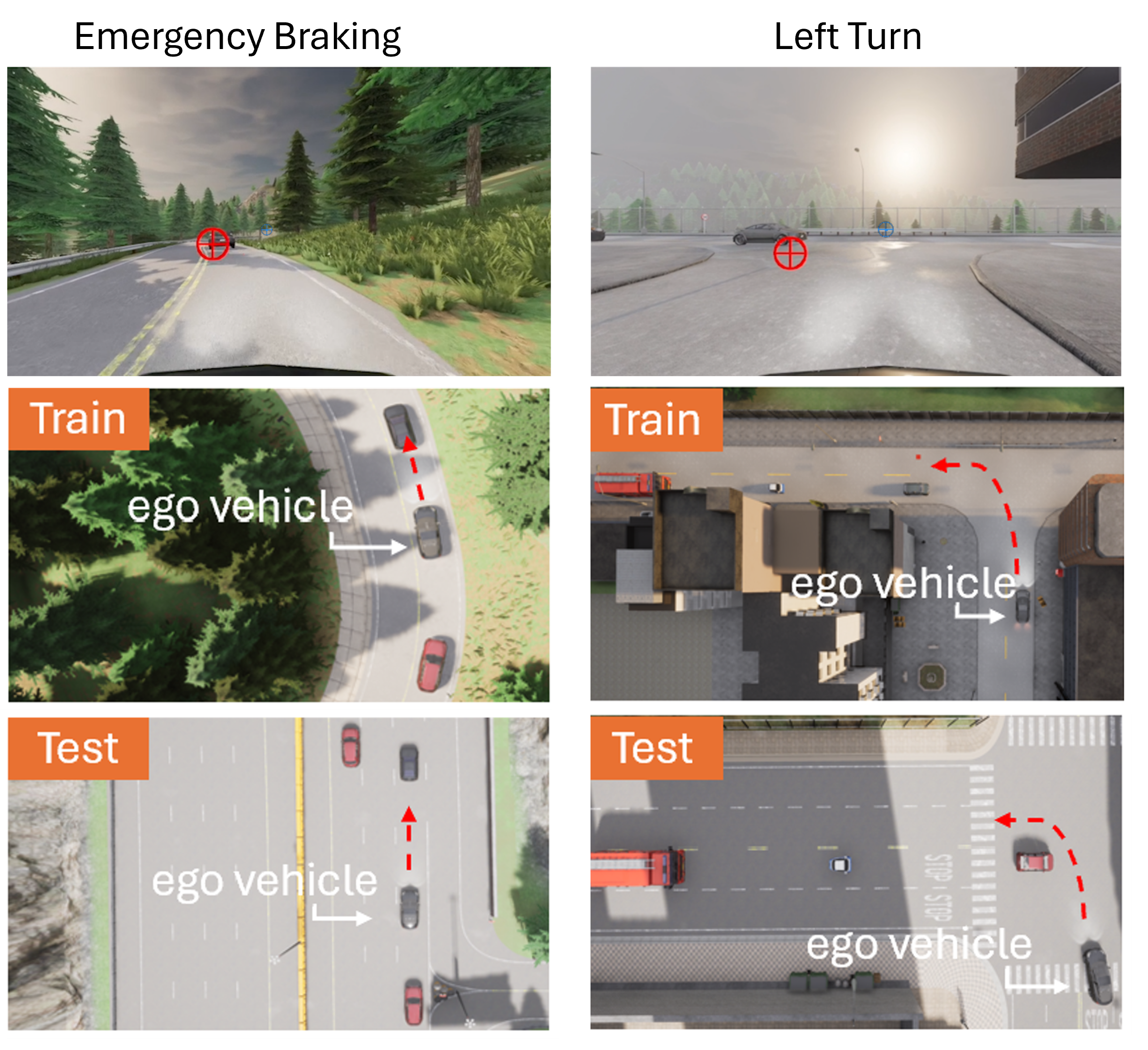}%
  \label{fig:sub-second}
\caption{Upper: The dataset collection environment. The HTC VIVE Pro Eye VR headset and Logitech G923 Racing Wheel and Pedal give the subject a more realistic driving experience. Lower: the example of scene image of driver's view, camera's view, and bird's eye view.}
\label{fig:supp_fig1}
\end{figure}

\section{Methods}

\subsection{EEG preprocessing}
\label{eeg_preprocessing}
The EEG preprocessing was done via the following pipeline in EEGLAB v2020.0 \cite{delorme2004eeglab} using MATLAB version 2022a. 
First, the EEG raw data were filtered using a bandpass filter with cutoff frequencies of 0.5 and 100 Hz, employing a zero-phase digital filter.
Second, the EEG data was downsampled from 1000 Hz to 250 Hz.
Next, the 50-Hz power line noise was removed via the CleanLine EEGLAB plugin using multi-tapering. 
Next, automated artifact rejection was applied via EEGLAB CleanArtifacts plugin. Bad channels were removed based on their standard deviations and correlation with other channels. 
 Bad portions of data were rejected with the Artifact Subspace Reconstruction (ASR) algorithm \cite{blum2019riemannian}. 
 Next, we replace the removed channels by interpolating them from the neighborhood channels. Independent Component Analysis (ICA) was then performed using the AMICA function, and dipole fitting was applied to identify neural components.
 Lastly, ICA components classified as non-brain sources with a probability of brain component less than 50\% (e.g., muscle, eye, or electrical noise) were removed using ICLabel \cite{pion2019iclabel} and dipole fitting, with components exhibiting a residual variance above 15\% being discarded. To analyze the ERPs, we defined the onset of the front vehicle's braking as the event marker. We discard trials if the subject did not focus on the car as revealed by the eye-tracking.


\subsection{EEG feature prediction model}
Since EEG signals are usually absent during both training and inference, we develop an EEG feature prediction network that estimates EEG activations from scene images. We focus on event-related potentials (ERPs), identified as the most salient EEG feature. 
To classify ERP-inducing trials, we apply a moving average filter with a window size of 20 samples and a peak-to-peak threshold of 1.7 µV, which was selected to balance the training set, resulting in a 50/50 class split, and also corresponds to minimum ERP peak amplitudes reported in~\cite{shubhadarshan2024effect}. Based on this threshold, we partition the dataset into high-ERP and low-ERP trials and perform five-fold cross-validation.
A lightweight CNN is designed to predict whether a trial induces an ERP. 
The architecture has three layers of convolutional layers and an average pooling layer to reduce the number of parameters and avoid overfitting, as summarized in Fig.~\ref{fig:framework}. 
To train the model, we use the Binary Cross-Entropy (BCE) loss, given by:
\begin{equation}
    \label{bceloss}
    \mathcal{L}_{BCE} = - \frac{1}{N} \sum_{i=1}^{N} \left[ y_i \log (\hat{y}_i) + (1 - y_i) \log (1 - \hat{y}_i) \right]
\end{equation}
where \( y_i \) is the ground truth label (1 for high ERP, 0 for low ERP), and \( \hat{y}_i \) is the predicted probability.  
The model processes sequences of segmentation images and outputs a binary classification indicating ERP presence. Its lightweight design enables real-time inference, supporting efficient RL training.

\subsection{Reinforcement Learning Problem Formulation}
\label{RL_Def}

We frame our task as a goal-oriented collision-avoidance problem, where the ego vehicle must control its acceleration and braking to safely navigate a dynamic environment. This control task is modeled as a Markov Decision Process (MDP), formally defined by the tuple \( \{S, A, P, R\} \). At each timestep \( t \), the agent observes a state \( s_t \in S \), chooses an action \( a_t \in A \), and transitions to a new state \( s_{t+1} \in S \) based on the transition dynamics \( P \). The agent receives a scalar reward \( r_t \) determined by the reward function \( R(s, a) \). The objective of the reinforcement learning (RL) agent is to learn an optimal policy \( \pi \) that maximizes the expected discounted cumulative reward
\( \sum_{k=0}^{\infty} \gamma^k r_{t+k} \),
where \( \gamma \in (0,1) \) denotes the discount factor used to prioritize immediate rewards over distant ones.
\\
\textbf{State Space:}  
The state space \( S \) comprises temporal sequences of visual observations. Specifically, each state is defined as \(
S = \big\{ s_t \mid s_t = \{I_{t-2}, I_{t-1}, I_t\}, \, t \in \mathbb{N} \big\} \), where each \( I \in \mathbb{R}^{h \times w \times 1} \) represents a single-channel semantic segmentation frame captured by a front-facing camera with a 90° field of view, simulated within the CARLA environment~\cite{dosovitskiy2017carla}. These image sequences provide the temporal context necessary for decision-making.
\\
\textbf{Action Space:}  
The action space \( A \) consists of continuous scalar values representing longitudinal control inputs:
\(
A = \{ a_t \mid a_t \in [-1, 1], \, t \in \mathbb{N} \}
\)
where \( a_t = -1 \) corresponds to maximum braking, and \( a_t = 1 \) indicates full throttle.
\\
\textbf{Reward Function:}  
The agent is incentivized to avoid collisions and maintain safe following distances. The total reward at each timestep is composed of cognitive reward and environment reward:
\begin{align}
  r_t &= R(\cdot | s_t, a_t) \notag
      = \beta r_{cog}(s_t)
         + r_{collide} \cdot (s_t \in C_{collide})\\
         &+ \omega r_{idle} \cdot (s_t \in C_{idle}) + \delta r_{gap}(s_t)
\end{align}

where the term \( r_{cog}(s_t) \) represents the predicted cognitive reward associated with state \( s_t \), as estimated by the EEG feature predictor (\( \hat{y}_i \), defined in Equation~\ref{bceloss}). A negative weight \( \beta = -1 \) is applied to penalize states associated with increased cognitive load or perceived difficulty. \( C_{collide} \) and \( C_{idle} \) denote the collision and idle states, respectively. 
The environment reward components are defined as follows: A severe penalty \( r_{collide} = -100 \) is applied upon collision. A mild penalty \( r_{idle} = -1 \) is incurred when the ego vehicle's speed falls below 0.2 m/s. \( r_{gap} \) provides a time-gap-based reward that encourages maintaining an optimal following distance.

\subsection{Policy network}

\begin{figure}[htb!]
  \centering
  \includegraphics[width=\linewidth]{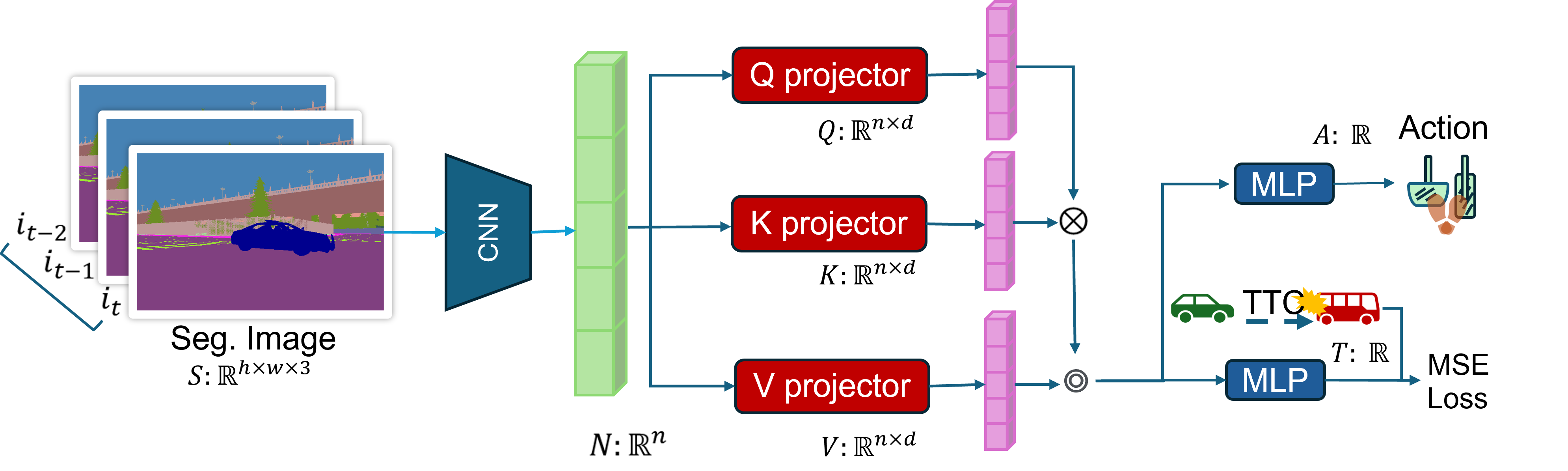}
  \caption{The policy network determines the vehicle's policy from a sequence of three segmentation images, starting with a CNN to extract features. These features are then processed through a self-attention layer. The policy network includes two MLP prediction heads: one for estimating the throttle and brake strength and another for predicting TTC, which aids in training regularization.
}
  \label{fig:policy_network}
\end{figure}

We design a policy network that incorporates a self-attention mechanism (see Fig.~\ref{fig:policy_network}).  
The model takes as input a sequence of three consecutive semantic segmentation images, forming the state representation  
\( s_t \in \mathbb{R}^{h \times w \times 3} \), where \( h \) and \( w \) denote the height and width of the images, respectively.  
The policy network jointly predicts two outputs within a unified framework: an action \( a_t \in [-1, 1] \) representing throttle and brake control, and the time-to-collision (TTC) in the range \( [0, 5] \).
A shallow CNN first encodes the input images into a feature map  
\( F \in \mathbb{R}^{h/16 \times w/16 \times f} \). A self-attention layer is applied to enlarge the receptive field as follows. The feature map is then flattened into  
\( N \in \mathbb{R}^{n \times f} \), where \( n = \frac{h}{16} \times \frac{w}{16} \), before being projected into query \( Q \), key \( K \),  
and value \( V \) matrices via fully connected layers \( f_Q, f_K, \) and \( f_V \) by 
\(
Q = f_Q(N), K = f_K(N),  V = f_V(N)
\), where \( Q, K, V \in \mathbb{R}^{n \times d} \), and \( d \) represents the dimensionality of the latent space for each token.  
These matrices are used to compute self-attention as follows: \(
\text{SelfAttention} = \text{softmax} \left( \frac{QK^\top}{\sqrt{d}} \right) V
\).
In this formulation, \( QK^\top \in \mathbb{R}^{n \times n} \) represents the pairwise attention scores between elements,  
which are scaled by \( \sqrt{d} \) to enhance numerical stability. The softmax function normalizes these scores,  
converting them into a probability distribution where each row sums to 1. 
These attention weights are applied to \( V \), producing the final contextualized feature representation. The output of the self-attention layer is then flattened and passed through multilayer perceptrons (MLPs) 
to predict the action \( a_t \in [-1, 1] \) and TTC, where \( -1 \) corresponds to maximum braking and \( 1 \) represents maximum throttle.  
The TTC prediction is computed as:
\begin{equation}
  TTC = \text{clip} \left({Dis} / (V_{ego} - V_{front}), 0, 5 \right)
  \label{eq:ttc}
\end{equation}
where \( Dis \) is the distance to the closest vehicle, \( V_{ego} \) is the ego vehicle's speed,  
and \( V_{front} \) is the speed of the nearest vehicle ahead.  
We clip \( TTC \) to the range \( [0,5] \) seconds to ensure the model focuses on critical situations.  
The mean squared error (MSE) loss \( \mathcal{L}_{mse} \) aligns the predicted TTC with the ground truth.
The overall loss function of the policy network is:
\(
  \mathcal{L}_{total} = \mathcal{L}_{\pi} + \alpha \mathcal{L}_{mse}
\),
where
\( \mathcal{L}_{\pi} \) is the RL policy loss,
 \( \mathcal{L}_{mse} \) is an auxiliary loss that regularizes TTC prediction.
\( \alpha = 0.1\) is a weighting factor that balances the RL loss and the MSE loss.
\section{Results}





\begin{figure}
\centerline{\includegraphics[width=18.5pc]{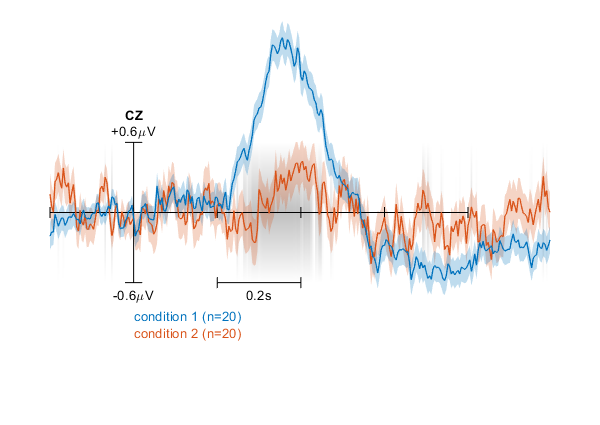}}
\caption{The average ERP wave from 20 participants. Condition 1: participants actively react to the emergency braking. Condition 2: participants are not required to react actively. The gray region indicates a significant difference between conditions.}
\label{fig:erp_cz}
\end{figure}


\begin{figure}[htb!]

  \centering
    \includegraphics[width=\linewidth]{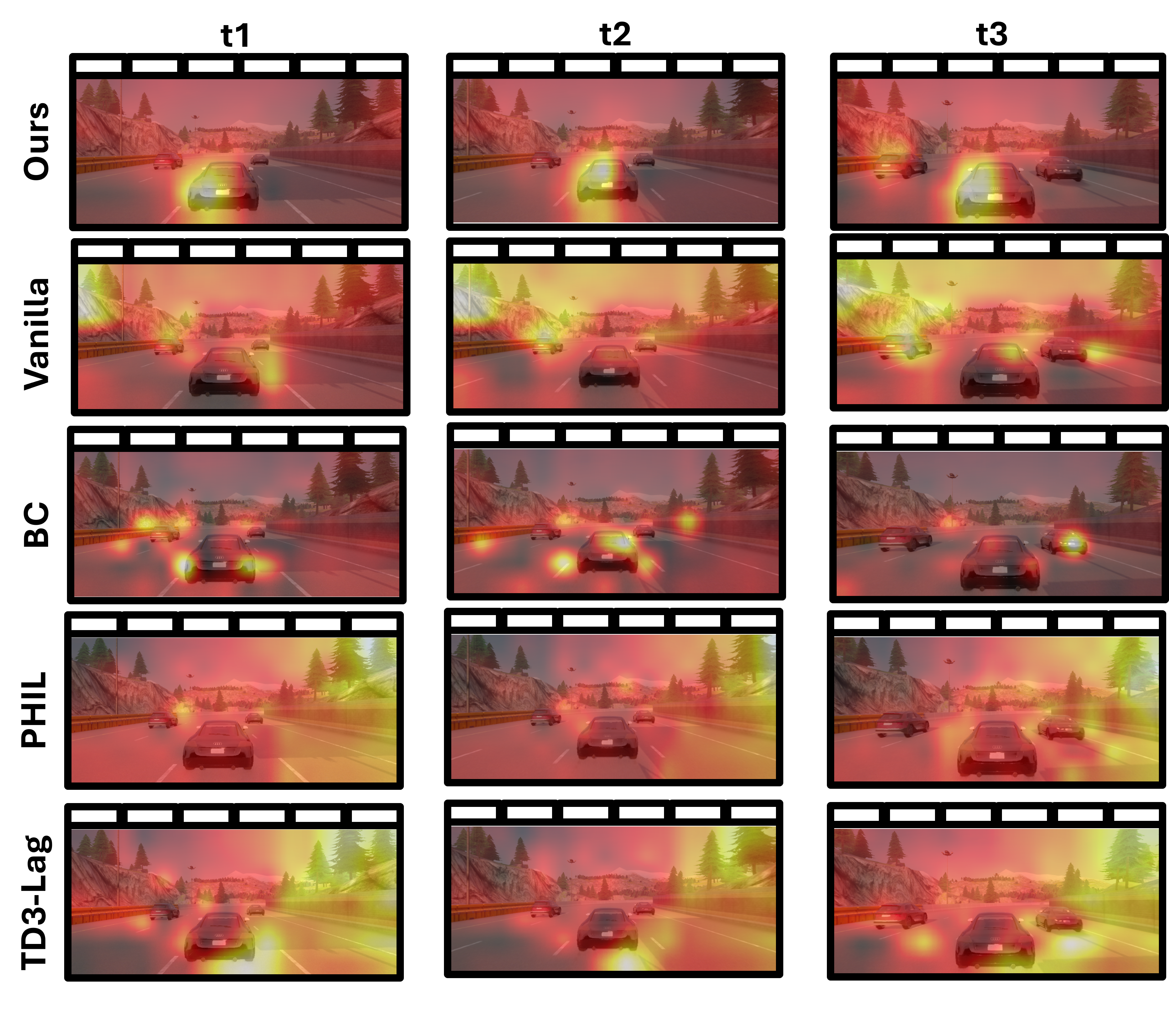}
    \label{supp_fig:information_sheet_1}
  \caption{Machine attention visualization of the policy network of RL across three time steps in the emergency braking scenario. Our model consistently focuses on the lead vehicle.}
  \label{supp_fig:policy_att}
\end{figure}



\subsection{EEG analysis}
We further investigated whether the ERP peak correlated with reaction time, which was defined as the interval between the front vehicle's braking and the subject's initiation of braking. The Pearson correlation between ERP peak latency and reaction time yielded a statistically significant result with p = 0.0438, indicating a significant positive association.
Additionally, we compared the ERP waveform under active reaction conditions to a baseline condition in which the subjects were not required to respond. The ERP amplitude was substantially higher when subjects responded to the emergency braking of the front vehicle, as shown in Fig.~\ref{fig:erp_cz}. To assess statistical significance, we conducted a non-parametric permutation test with 10,000 iterations. The shaded region between 300 and 500 ms revealed a significant difference. This finding aligns with our expectations, as the P3 amplitude is known to increase with greater cognitive effort, supporting the interpretation that P3 amplitude may serve as an index of resource allocation~\cite{isreal1980p300}.

\subsection{EEG feature prediction model}
To eliminate the need for EEG data collection during inference, we propose a human cognitive reward model that predicts whether the trial will induce ERP directly from visual input. To ensure robustness across subjects, we conducted five-fold cross-validation. We first compared our EEG feature prediction model with several commonly used image classification baseline methods, including ResNet-18 \cite{he2016deep}, Swin-ViT~\cite{liu2021swin}, and ConNeXt~\cite{liu2022convnet}. Our model achieved a slightly higher average accuracy of 82\%.  Notably, it reached a processing speed of 204 frames per second (fps), making it more suitable for use as a pretrained model within the RLHF framework. When training the RL with 1M steps, it can save 2.1 hours of training time compared with Resnet-18.

\begin{table}[h]

\centering
\begin{tabular}{@{}lccccccc@{}}
\toprule
      & $\mathbf{F_1}$ & $\mathbf{F_2}$ & $\mathbf{F_3}$ & $\mathbf{F_4}$ & $\mathbf{F_5}$ & \textbf{Mean} & \textbf{FPS} \\ \midrule
Resnet-18 & 82  & 89 & 79 & 79 & 76 & 81 & 73 \\ 
Swin-ViT & 80  & 85 & 75 & 80 & 77 & 79 & 62\\
ConvNeXt & 82  & 89 & 75 & 81 & 76 & 80 & 74 \\
Ours & 80  & 85 & 77 & 86 & 81 & 82 & 204\\ \bottomrule
\end{tabular}

\caption{Five-fold cross-validation result and the FPS of the EEG feature prediction model.}
\end{table}


\subsection{Driving performance analysis}

\subsubsection{Baseline methods}
\textbf{Vanilla RL: }
For reinforcement learning, we adopt the TD3 algorithm \cite{fujimoto2018addressing}, an off-policy approach known for its robustness and improved sample efficiency in continuous control tasks.
\textbf{Behavior Cloning (BC): }
In addition to using the standard TD3 algorithm, we include a baseline that leverages BC \cite{bain1995framework} to initialize the policy. Specifically, we construct a supervised learning model trained on human driving demonstrations, using the architecture shown in Fig.~\ref{fig:policy_network}. The model is optimized using an MSE loss.
 \begin{table*}[htb!]

\centering
\begin{tabular}{@{}lccc ccc@{}}
\toprule
\multirow{2}{*} &
\multicolumn{3}{c}{Emergency Braking} &
\multicolumn{3}{c}{Left Turn} \\
\cmidrule(lr){2-4} \cmidrule(lr){5-7}
& Route Completion$\uparrow$ & Driving Score$\uparrow$ & Infraction$\uparrow$
&Route Completion$\uparrow$ & Driving Score$\uparrow$ & Infraction$\uparrow$ \\
\midrule
Vanilla & $23 \pm 27$ & $16 \pm 19$ & $0.66 \pm 0.07$ & $60 \pm 16$ & $45 \pm 19$ & $0.68 \pm 0.09$\\
BC      & $65 \pm 31$ & $55 \pm 29$ & $0.72\pm 0.08$ & $48 \pm 5$ & $29 \pm 3$ &  $	0.62 \pm 0.04$ \\
PHIL    & $59 \pm 23$ & $44 \pm 29$ & $0.67 \pm 0.14$ & $38 \pm 32$ & $31 \pm 30$ & $0.68 \pm 0.11$\\
RLHF & $73 \pm 32$ & $66 \pm 39$ & $0.80 \pm 0.19$ & $63 \pm 21$ & $49 \pm 28$ & $0.71 \pm 0.15$\\  
TD3-lag     & $44 \pm 33$ & $35 \pm 34$ & $0.72 \pm 0.16$ & $40 \pm 28$ & $32 \pm 22$ & $0.68 \pm 0.06$ \\
\midrule
Ours      & $85 \pm 43$ & $79 \pm 31$ & $0.84 \pm 0.14$ & $67 \pm 8$ & $57 \pm 10$ & $0.77 \pm 0.05$\\ \bottomrule
\end{tabular}

\caption{Summary statistics for Route Completion, Driving Scores, Infraction Scores for emergency braking and left-turn scenario.}
\label{tab:main_ds}
\end{table*}
\textbf{Prioritized Human In-the-Loop (PHIL) RL: }
 PHIL utilized the Prioritized Experience Replay (PER) \cite{schaul2015prioritized} mechanism, which organizes the replay buffer based on the magnitude of Temporal Difference (TD) errors. PHIL replaces the exploratory experience of RL with human guidance into the experience replay buffer, and prioritized experience with a high TD error. 
\textbf{TD3-lag: }
TD3-lag is a safe RL method based on TD3, augmented with a Lagrangian-based approach to model collision as a cost function to enforce safety constraints~\cite{ji2023safety}. 
\textbf{RLHF: } A custom interface was developed to present side-by-side pairs of clips sampled from our RL agents exhibiting diverse driving behaviors, as shown in supplementary materials. Each video takes 2 seconds following the RLHF framework~\cite{christiano2017deep}. Three participants completed 2000 pairwise queries, each selecting one of three options: “prefer left,” “prefer right,” or “cannot tell.” The whole data collection takes about 10 hours of manual labelling. The responses were stored in csv format and used to train a human preference reward model with the Bradley-Terry loss~\cite{christiano2017deep}. This reward model was then integrated into RL training under identical conditions to our ERP-based predictor.

\subsubsection{Evaluation metrics}

We used three commonly used evaluation metrics in CARLA, including \textit{driving score}, \textit{route completion}, and \textit{infraction score} as our evaluation metrics. 
\\
\textbf{Driving Score:} The driving score is defined as a weighted sum of the product of route completion and infraction penalty: \(
\text{Driving Score} = \sum_i w_i \cdot (R_i \cdot P_i)
\). Let $R_i$ be the percentage of completion of the $i$-th route, and $P_i$ the infraction penalty of the $i$-th route. The driving score has a maximum value of 100.
\textbf{Route Completion ($R_i$):} Percentage of the route distance completed by an agent before a collision or reaching the destination. It has a maximum value of 100.
\textbf{Infraction Penalty ($P_i$):}
Each collision will introduce a discount factor of 0.6. Agents start with an ideal base score of $1$, which is multiplicatively reduced each time an infraction is committed, down to a minimum value of $0$.

\subsubsection{Evaluation results}

To ensure the robustness and reproducibility of our findings, we trained five models using different random seeds. 
To evaluate the generalization capability of the RL agents, we trained and evaluated them in different towns. Specifically, we trained the agents in CARLA town 7 and evaluated them in town 4 for the emergency braking scenarios, and trained in town 1 and evaluated in town 5 for the left-turn scenario. Our method achieved the highest route completion score of 85 and 67, infraction penalty score of 0.84 and 0.77, and the driving score of 79 and 57, showing the effectiveness of our framework.
We investigate whether the cognitive reward model enhances the internal representation of the policy network by visualizing its machine attention. Specifically, we examine what the policy network has learned through the self-attention mechanism, computed as follows: 
\(
\text{MachineAttention} = \text{softmax}\left(\frac{QK^\top}{\sqrt{d}}\right).
\)
The resulting machine attention matrix is averaged across all rows to produce a mean attention vector, which is subsequently reshaped into a \(h/16 \times w/16\) feature map, reflecting the spatial distribution of aggregated attention over the 2D layout.
As illustrated in Fig.~\ref{supp_fig:policy_att}, our model consistently attends to the lead vehicle across all time steps, whereas the machine attention in other baseline models appears more dispersed.
\section{Limitations}

This study includes only two scenarios involving 20 participants, as some individuals experienced motion sickness, limiting broader participation. In future work, the integration of foveated rendering technology may help mitigate motion sickness, enabling more diverse and extensive data collection.
Additionally, our current EEG feature prediction model is scenario-specific. Nevertheless, it serves as a foundational step toward leveraging EEG-based feature prediction as implicit feedback for RL agents. With the low-cost data collection system proposed in this work, there is potential to train a more generalizable model in the future as larger and more varied datasets become available.
\section{Conclusion}
\label{sec:conclusion}

In this work, we present an RL framework that integrates human cognitive reward feedback derived from EEG signals to enhance intelligent vehicle performance. To the best of our knowledge, this is the first study to introduce EEG-based reward modeling in the context of autonomous driving.
First, we collected a multimodal dataset from 20 participants engaged in a simulated driving task with eye-tracking, EEG, and control data. ERP analysis revealed a prominent P3 component.
To eliminate the need for collecting EEG data during inference, we developed a lightweight EEG feature prediction model capable of estimating ERP occurrence from scene images.
Finally, we integrated the predicted cognitive signals into an RL framework and evaluated our approach in the CARLA driving simulator. The results demonstrate that agents trained with cognitive reward feedback exhibit safer driving behavior.
Overall, our findings suggest that human cognitive signals, even when indirectly inferred, can provide valuable guidance for training safer and more human-aligned autonomous systems.

\section{Acknowledgement}
\label{sec:acknowledgement}

This work was supported in part by the Australian Research Council (ARC) under discovery grant DP250103612 and DP260101395, ARC Research Hub for Human-Robot Teaming for Sustainable and Resilient Construction (ITRH) grant IH240100016, and Australian National Health and Medical Research Council (NHMRC) Ideas Grant APP2021183. Research was also sponsored in part by the Australia Advanced Strategic Capabilities Accelerator (ASCA) under Contract No. P18-650825 and ASCA EDT DA ID12994, and the Australian Defence Science Technology Group (DSTG) under Agreement No: 12549. Last but not least, many thanks to Mr. Cheng-you Lu for the insightful discussion and Mrs. Haiting Lan for proofreading.

{
    \small
    \bibliographystyle{ieeenat_fullname}
    \bibliography{main}
}


\end{document}